\documentclass{article}
\usepackage{ijcai24}

\usepackage{times}
\usepackage{soul}
\usepackage{url}
\usepackage[hidelinks]{hyperref}
\usepackage[utf8]{inputenc}
\usepackage[small]{caption}
\usepackage{multirow}
\usepackage{graphicx}
\usepackage{amsmath}
\usepackage{amsthm}
\usepackage{subfigure}
\usepackage{epsfig}
\usepackage{amssymb}
\usepackage{pifont}
\usepackage{booktabs}
\usepackage{algorithm}
\usepackage{algorithmic}
\usepackage{lipsum}
\usepackage{dsfont}
\usepackage[table]{xcolor}
\usepackage[switch]{lineno}
\usepackage[capitalize]{cleveref}

\title{OD-DETR: Online Distillation for Stabilizing Training of Detection Transformer}

\author{
Shengjian Wu$^{1,2}$
\and
Li Sun$^{1,3*}$\and
Qingli Li$^1$\\
\affiliations
$^1$Shanghai Key Laboratory of Multidimensional Information Processing, East China Normal University\\
$^2$FinVolution Group\\
$^3$Key Laboratory of Advanced Theory and Application in Statistics and Data Science, East China Normal University\\
\emails
sunli@ee.ecnu.edu.cn
}

\begin{document}

\maketitle

\begin{abstract}
    DEtection TRansformer (DETR) becomes a dominant paradigm, 
    mainly due to its common architecture with high accuracy and no post-processing. However, DETR suffers from unstable training dynamics. It consumes more data and epochs to converge compared with CNN-based detectors. This paper aims to stabilize DETR training through the online distillation. 
    It utilizes a teacher model, accumulated by Exponential Moving Average (EMA), and distills its knowledge into the online model 
    in following three aspects. 
    \emph{First}, 
    the matching relation between object queries 
    and 
    ground truth (GT) boxes in the teacher 
    is employed to guide the student, 
    so 
    queries within the student are not only 
    assigned labels based on their own predictions, but also refer to the matching results from the teacher. \emph{Second}, the teacher's initial query is given to the online student, and its prediction is directly constrained by the corresponding output from the teacher. \emph{Finally}, the object queries 
    from teacher's 
    different decoder stages 
    are used to build the auxiliary group to accelerate the convergence. For each GT, two queries with the least matching costs are selected into this extra group, and they predict the GT box and participate the 
    optimization. Extensive experiments show that the proposed OD-DETR successfully stabilizes the training, and significantly increases the performance without bringing in more parameters.
\end{abstract}

\begin{figure}[!htb]
    \centering
    \includegraphics[width=7cm]{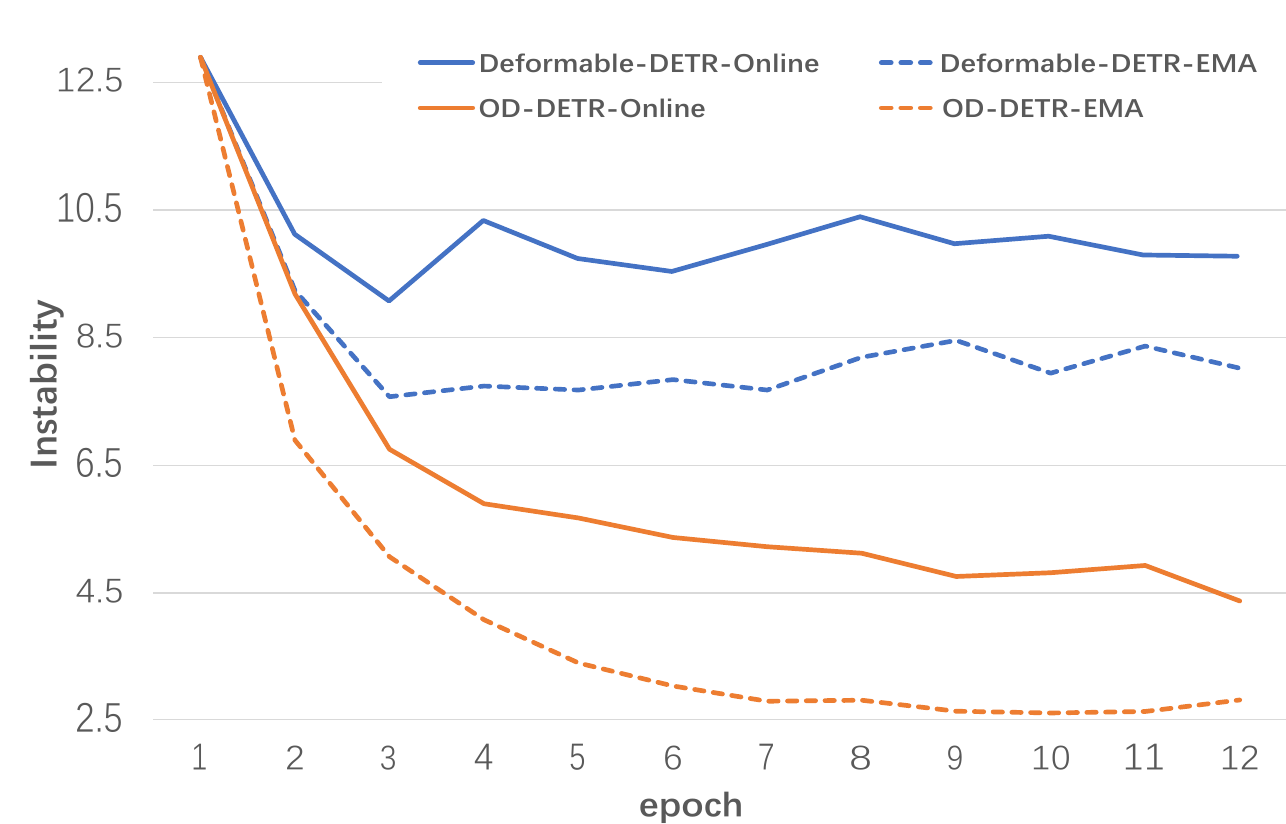}
    \caption{ 
    The matching instability curves for the first 12 training epochs. We compare our OD-DETR and its EMA version with Def-DETR. 
    The metric, introduced by \protect\cite{li2022dn}, is calculated on COCO VAL2017. Lower value 
    means more stable matching. As is expected, EMA's 
    instability is much lower 
    than the online model. In OD-DETR, the online student learns from its 
    EMA teacher, 
    which greatly increases its 
    stability. 
    The improved online model also helps to stabilize the EMA's matching results. 
    }
    \label{fig:instability}
\end{figure}

\newcommand\blfootnote[1]{%
\begingroup
\renewcommand\thefootnote{}\footnote{#1}%
\addtocounter{footnote}{-1}%
\endgroup
}

\blfootnote{*Corresponding Author}

\begin{figure*}[htb]
    \centering
    \includegraphics[width=16.5cm]{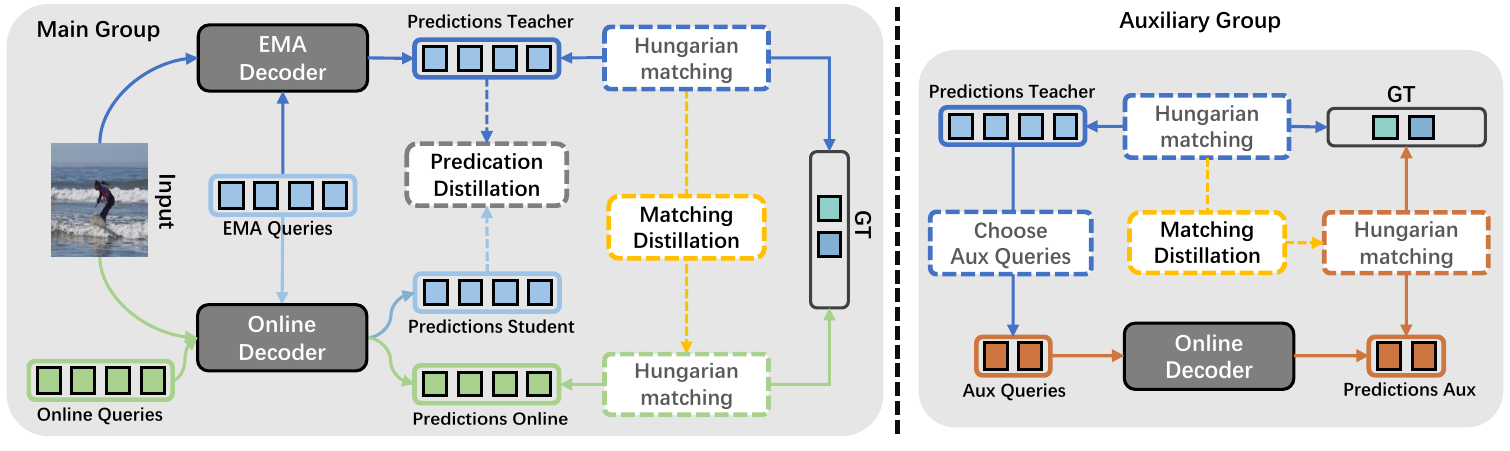}
    \caption{The overall architecture of OD-DETR. 
In the Main Group, 
the EMA model's queries are 
given to 
the decoder to create Predictions Teacher, and they are matched with the GT set using Hungarian matching. Simultaneously, these queries are also input into 
the Online Decoder to produce Predictions Student, 
and they directly learn from Predictions Teacher through prediction distillation. The Online model's own queries are 
decoded into Predictions Online, which are also 
matched with the GT set. 
Then, through matching distillation, 
they refer to matching result of Predictions Teacher. 
In the Auxiliary Group, 
we select two updated queries with the lowest matching cost for each GT from the Predictions Teacher.
These selections are added as an extra 
group to the Online Decoder. 
}
    \label{fig:pipeline}
\end{figure*}

\section{Introduction}
Object detection is a fundamental task in computer vision, and has been investigated by the community for decades. CNN-based detectors can be categorized into either anchor-based \cite{7410526,liu2016ssd} or anchor-free \cite{tian2019fcos,Yang2019RepPointsPS} methods. The former builds upon the sliding anchor boxes, and can be designed into a single, two or multi-stage, 
while the latter only has grid point assumption and is generally of a single stage. Although, CNN-based detectors has achieved impressive performance, it needs to determine complex meta parameters like anchor shape and size, threshold 
for positive and negative samples, and non-maximum suppression (NMS) post-processing.

DEtection TRansformer (DETR) greatly simplifies the cumbersome design. It employs the encoder attention 
to enhance image feature. At the same time, several decoder attention layers are also incorporated, which translate the initial parameters of learnable object queries into predicted 
boxes. DETR uses bipartite matching to setup the one-to-one relation between ground truths (GTs) and predictions 
from queries, therefore, one GT is assigned to a single query, and vice versa. The one-to-one matching scheme reduces the redundant predictions, and alleviates detector from NMS. 
However, DETR is often blamed for its unstable training and slow convergence. As is shown in \cref{fig:instability}, the GT for query is often switched during training. Many efforts intend to improve it, by either introducing local box prior \cite{meng2021conditional,zhu2020deformable,liu2022dab}, 
more query groups \cite{li2022dn,chen2022group,jia2022detrs}, an initial stage \cite{Yao2021EfficientDI,zhang2022dino} or improved quality-aware loss functions \cite{liu2023detection,cai2023aligndetr}.

This paper proposes a solution, named online distillation (OD-DETR), in another perspective to 
stabilizing the training of DETR. Inspired by the 
works in semi-supervised classification \cite{sohn2020fixmatch} and 
object detection \cite{liu2021unbiased}, we 
utilize Exponential Moving Average (EMA) 
as the teacher and distill its knowledge into the 
student model in an online manner. Unlike traditional distillation, the teacher also gets improved 
by the accumulation 
of the student. Particularly, we leverage the 
predicted bounding boxes, Hungarian matching result and updated object queries from the teacher model, and design schemes for prediction distillation, matching distillation and building auxiliary group within the student. For matching distillation, we 
assign another label for each object query 
through Hungarian matching according to cost matrix between teacher's predictions and GT boxes, and use it to guide the online student together with its original matching results. 
To employ two possible matched GTs, 
we propose a multi-target QFL 
loss to accommodate two labels from different classes, while keep only one 
regression target to avoid ambiguity. 
Meanwhile, two predictions associated with the same GT are given different regression loss weights, and the one with larger matching cost is down-weighted. 

To 
take full advantage of the teacher, 
its initial queries 
are fed into the online student, giving predictions 
that can be directly constrained by the corresponding output from the teacher with no need to rematch. We name this simple 
constraint between two bounding boxes as prediction distillation. To further strengthen the link between teacher and student, we build independent augmented groups by object queries from each decoding stage of the teacher. Here, high quality queries with minimal matching costs for each GT are selected, and 
each group 
is given to the student's decoder to make predictions, re-match with GTs and compute losses. The augmented query groups are mainly used for speeding up the training convergence, hence it is abandoned during inference. The framework of OD-DETR is given in \cref{fig:pipeline}. 

To verify the effectiveness of OD-DETR, we perform a large amount of experiments on MS-COCO \cite{lin2014microsoft}. Particularly, 
OD-DETR is implemented to support different variations of DETR including Def-, DAB- and DINO. We find that our method is compatible with all of them and increases the performance obviously. In summary, the contributions of this paper lie in following aspects.
\begin{itemize}
    \item We propose an EMA-teacher based online distillation scheme. 
    Particularly, matching distillation between the teacher and student is proposed. 
    It assigns each query to an extra GT box based on the matching results of teacher's prediction. Moreover, we adapt the loss functions and propose a multi-target QFL classification loss and a cost sensitive regression loss.
    \item We perform prediction distillation and build augmented query groups to 
    fully utilize of the EMA teacher. 
    Prediction distillation is carried out by feeding the EMA query into student's decoder, and constraining the prediction by teacher's output, while the augmented groups are built by high-quality queries from different decoding stages within the teacher.
    \item Intensive experiments are carried out, showing that the proposed OD-DETR is able to effectively increase the performance of different DETR's variations.
\end{itemize}

\section{Related Work}

\subsection{Supervised Object Detection}

Modern object detection models mainly use convolutional networks and achieved great success recently. These CNN-based detectors fall into two categories: anchor-based and anchor-free. \cite{7410526,Ren2015FasterRT,lin2017focal} are some well-known anchor-based model, while \cite{tian2019fcos,Yang2019RepPointsPS,Duan2019CenterNetKT} represent anchor-free models. Both both of them need manual components like non-maximum suppression (NMS) and heuristic label assignment rules.

DEtection TRansformer (DETR) \cite{carion2020end} changes 
this scenario. It's the first end-to-end, query-based object detector without handcrafted 
designs such as anchors and NMS. However, it suffers from the slow training convergence. 
Several recent studies have focused on speeding up DETR's training process. 
Methods such as \cite{zhu2020deformable,gao2021fast,gao2021fast,Sun2020SparseRE,Zhao2022IoUEnhancedAF,Zhao2023RecursiveDetER,liu2022dab} employ local box priors to concentrate on local features, reducing the search space.
Other methods like \cite{chen2022group,jia2022detrs,li2022dn,zhang2022dino}
speed up training by adding more query groups, providing additional positive samples for ground truth boxes.

\subsection{EMA in SSOD and Distillation}
Using 
EMA model as teacher to distill knowledge into student 
has been consistently applied in semi-supervised learning (SSL) based classifications, 
examples of which include \cite{zhu2005semi,laine2016temporal,tarvainen2017mean,berthelot2019mixmatch,sohn2020fixmatch}. A critical challenge in SSL is to fully leverage unlabeled images, which is 
realized 
by either self-training or consistency regularization.
Another area where EMA-based teacher approaches is applied 
is 
in self-supervised learning, as exemplified by methods like \cite{Grill2020BootstrapYO,Gidaris2020OBoWOB,Caron2021EmergingPI}. The key to self-supervised learning also lies in the student model learning from the teacher model's outputs on unlabeled data.

Based on the similar idea in classification task, 
numerous SSOD (Semi-Supervised Object Detection) models have been developed. For instance, \cite{tang2021humble,xu2021end,liu2021unbiased} adopt the principles of FixMatch \cite{sohn2020fixmatch} and employs an EMA accumulated from the online student to provide pseudo GT boxes. 
However, for knowledge distillation (KD) task in object detection, current methods typically use a fixed, pre-trained model as the teacher to output labels, which are then provided to the online student for learning, as can be seen in 
\cite{NIPS2017_e1e32e23,8100259,Chang2022DETRDistillAU,Chen2022D3ETRDD,Wang2022KnowledgeDF}. Different from works in SSOD or KD, this paper focuses on online supervised learning for detection, and we utilize the EMA model as the teacher to provide guidance 
in an online manner, without requiring a fixed teacher model optimized in advance.

\begin{figure*}[ht]
    \centering
    \includegraphics[width=14.7cm]{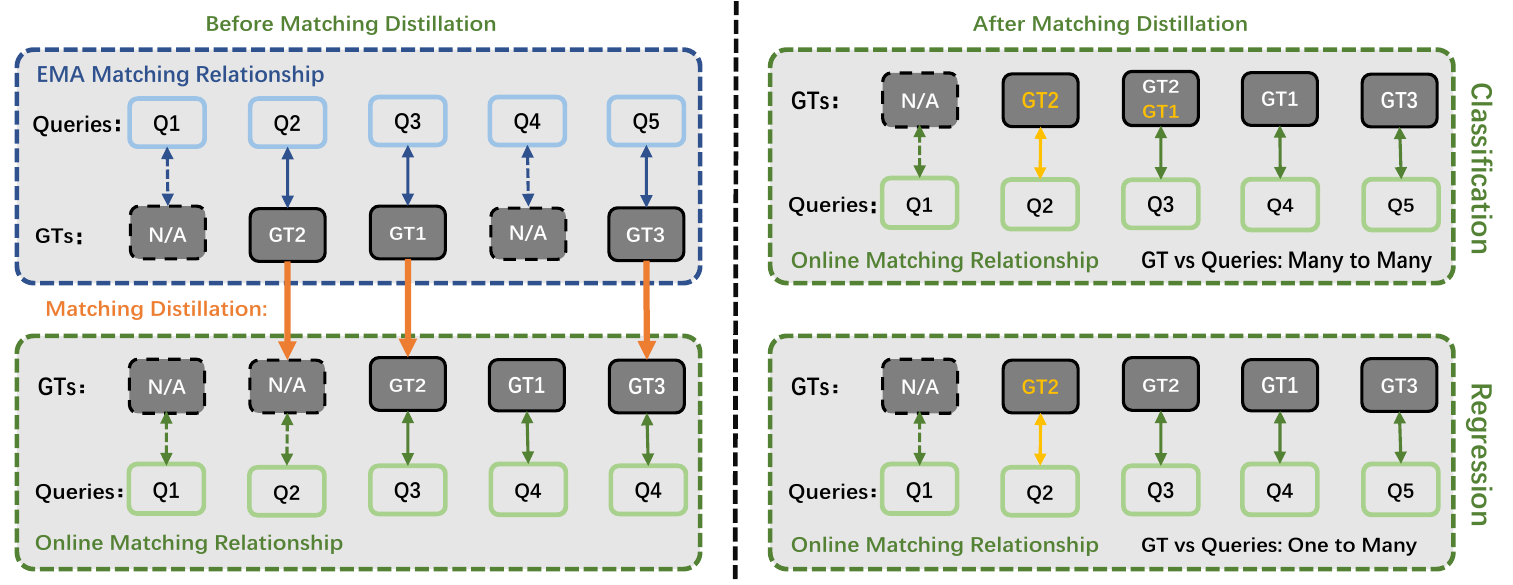}
    \caption{Matching Distillation 
    injects GTs matched in the EMA teacher into the corresponding queries in the online student. On the left, two independent matching are first carried out, assigning GTs for queries in the EMA and online models, respectively. 
    On the right, 
    new matches for online model are shown in gold color. 
    For classification, this creates many-to-many matches, where queries like Q3 can match with GT2 and GT1. But for regression, to avoid confusion, each query 
    is assigned with one GT, making a one-to-many match between GTs and queries. }
    \label{fig:MD_sample}
\end{figure*}

\section{Method}
\subsection{Preliminaries on DETR and Its Adaptation 
}
DETR is composed of a backbone, an encoder of several self-attention layers, a set of learnable object queries, and a decoder 
followed by a detection head to translate updated queries into boxes with class predictions. 
We define 
$\mathcal{Q}=\{q_i|q_i\in\mathbb{R}^c\}$ as the query set. Each $q_i$ is a learnable parameters at the beginning stage. 
To make the model aware of the positions of bounding boxes, the enhanced versions of DETR \cite{liu2022dab,zhu2020deformable} explicitly encode the bounding box 
$(x,y,w,h)$ or only box center $(x,y)$ into positional embedding, specifying a set $\mathcal{P}=\{p_i|p_i\in\mathbb{R}^c\}$ corresponding to $\mathcal{Q}$. $\mathcal{P}$, $\mathcal{Q}$ and image feature $\mathcal{F}$ are given to the decoder $Dec$, resulting in an updated query set $\mathcal{Q}$ and box set $\mathcal{B}$ for the next stage.
\begin{equation}\label{eq:eq1}
    [\mathcal{Q}^{t+1},\mathcal{B}^{t+1}] = Dec^t(\mathcal{Q}^t,\mathcal{P}^t,\mathcal{F};\theta)
\end{equation}
In \cref{eq:eq1}, $Dec$ indicates the decoder parameterized by $\theta$, and the superscript denotes the decoder stage. The predicted box set $\mathcal{B}=\{b_i|b_i=(x,y,w,h,c)\}$ not only has box coordinate but also a class score vector $c$. 
Apart from the detection head, the DETR's decoder 
consists of self attention layer computed by interactions among query elements $q_i$ in $\mathcal{Q}$, cross attention layer interacting between $\mathcal{Q}$ and $\mathcal{F}$, and feed-forward network between them. In Def-DETR \cite{zhu2020deformable}, the cross attention can be improved by only sampling 
features around the reference point and giving a weighted average, and this local prior speeds up its convergence. 

The updated $\mathcal{Q}$ are given to the detection head for classification and bounding box regression. 
But before 
loss computation, the predictions 
from 
queries are assigned to GTs 
according to matching costs. 
DETR adopts the Hungarian algorithm 
to set up the one-to-one relation between query set $\mathcal{Q}$ and all GTs, which is also the key reason that DETR can be free from NMS. However, this dynamic one-to-one matching strategy also leads to training instability. The work \cite{li2022dn} finds that there are some queries assigned with different GTs between two training epochs, as is shown in the blue curve in \cref{fig:instability}, and it causes the slow convergence of DETR.

The training target of original DETR includes a focal loss $L_{cls}$ \cite{lin2017focal} for classification, and $L_1$ and GIoU loss $L_{GIoU}$ for regression. Some works show that quality metric can improve performance. Notably, with quality focal loss $L_{QFL}$ \cite{li2020generalized} defined in \cref{eq:eq2}, classification and regression tasks are bounded together. Here $t$ is the IoU target calculated between predicted box $b$ and its matched GT box. $s$ is a prediction 
score from sigmoid function. 
\begin{equation}\label{eq:eq2}
L_{QFL}(s,t)=-|t-s|^\gamma((1-t)\log(1-s)+t\log(s))
\end{equation}

Our proposed OD-DETR is built on enhanced DETR with QFL classification loss. Next, we introduce three key components: matching distillation, prediction distillation, and auxiliary group, as is shown in Fig.\ref{fig:pipeline}.

\begin{figure}[!hb]
    \centering
    \includegraphics[width=6.5cm]{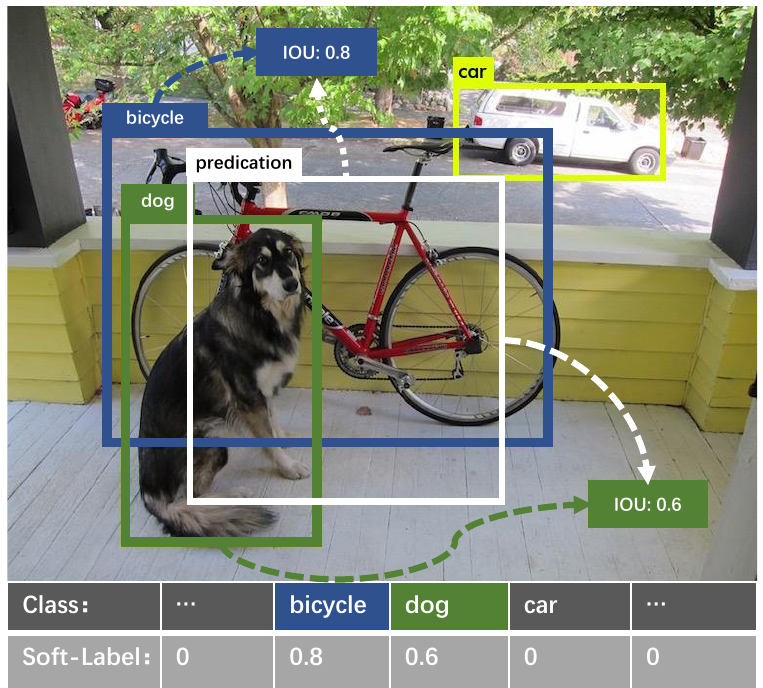}
    \caption{Illustration for setting the multi-target label. The prediction shown here matches two objects of different classes, the dog and the bicycle, with IOUs of 0.6 and 0.8, respectively. In its label vector, 
    two entries for dog and bicycle are set accordingly. 
    Other category elements stay at 0. 
    }
    \label{fig:soft_label}
\end{figure}

\subsection{Matching Distillation}
Inspired by the success of the EMA model, we 
take advantage of it to stabilize DETR training. 
We first validate 
its potential application in Def-DETR. Specifically, we train 
it in 12 epochs and make an EMA at the same time. 
We compare it with the online 
model based on the instability 
metrics defined in \cite{li2022dn}. As is shown 
in \cref{fig:instability}, 
the behaviour of EMA 
is 
more stable than the online model, 
showing that it indeed prevents the label switch between two epochs, therefore giving a stable matching result. 
Considering that the EMA model has a more stable matching result, 
we intend to distill it 
into the online student.

 Fig.\ref{fig:MD_sample} presents an 
 intuitive illustration of matching distillation. Matching result from the teacher and the student are combined into classification and regression losses for training the online model. 
Particularly, given teacher's query, PE and image feature $\mathcal{Q}'$, 
$\mathcal{P}'$ and $\mathcal{F}'$, the decoder of the teacher, parameterized by $\theta'$, can output predicted boxes $\mathcal{B}'$, as in 
\cref{eq:eq1}. Hungarian matching result 
between $\mathcal{B}'$ and the GT set 
is used as a reference for constraining the prediction $\mathcal{B}$ from student model. Note that 
$\mathcal{B}$ 
also 
have a matching result, which means that $q_i$ and its EMA version $q_i'$ may have different GT to be matched. We now illustrate matching distillation for classification and regression in the following two sections. 

\paragraph{Matching distillation for classification.} 
Since the online query $q_i$ 
are tightly connected with $q_i'$, we design a \textbf{multi-target QFL $L_{MQF}$} for class prediction in \cref{eq:eq4}, 
which utilizes the matched target label of $q_i'$, if it is different from the original target. Here $s'$ 
is another predicted class score, and $t'$ is its corresponding IoU target computed based on the teacher's matching result.
\begin{equation}\label{eq:eq4}\small
L_{MQF}(s,s',t,t')=
\begin{cases}
L_{QFL}(s,t) & \text{one class}\\
L_{QFL}(s,t)+L_{QFL}(s',t')& \text{two classes}
\end{cases}
\end{equation}
Particularly, if two matched GT are from the same class, $L_{MQF}$ becomes the same with $L_{QFL}$ defined in \cref{eq:eq2}, 
computed by its original IoU target $t$. Otherwise, if two GTs are in different classes, two IoU targets $t$ and $t'$ impose constraints on $s$ and $s'$, respectively. 
Fig.\ref{fig:soft_label} shows 
how the multi-target label is set for a query's prediction that matches two different categories of GTs. 
In dense scenes, 
one predicted box often contains multiple objects 
of different categories. 
Therefore, the single-category one-hot 
label 
is not appropriate. 
Our $L_{MQF}$ dynamically sets targets for different categories based on the IoU 
between the predicted box 
and two matched GTs, providing richer semantic information for training. It also 
increases training stability by preventing sudden label changes 
when online 
matching result changes.

\paragraph{Matching distillation for regression.} For bounding box regression, 
the matching result from the teacher are also referred to. But we avoid the ambiguity due to the case in which the two different GT boxes are matched with one query, and only use the online matched result as the target to compute $L_1$ and $L_{GIoU}$. 
Moreover, since one GT may still be matched by two queries, 
we 
simply \textbf{down-weight} the regression loss for the query with a bigger matching cost. The regression loss $L_{r}$ is defined in \cref{eq:eq5}. Here $b$ and $b_{gt}$ are the predicted and matched GT boxes. $w_d$ is a hyper-parameter with value of 0.51 following the idea in \cite{cai2023aligndetr}.
\begin{equation}\label{eq:eq5}\small
L_{r}(b,b_{gt})=
\begin{cases}
L_1(b,b_{gt})+L_{GIoU}(b,b_{gt})&b\text{ with lower cost}\\
w_d[L_1(b,b_{gt})+L_{GIoU}(b,b_{gt})]&b\text{ with higher cost}
\end{cases}
\end{equation}

\subsection{Prediction Distillation}
Besides the matching distillation, 
the output from the teacher model $\mathcal{B}'$ can also be exploited for training, 
and we name it prediction distillation. However, since 
the online predictions 
are already constrained by GT boxes, it is ambiguous to require them to approach another set of targets. To take full advantage of 
$\mathcal{B}'$, we feed the EMA queries $\mathcal{Q}'$ together with the PE $\mathcal{P}$ and image feature $\mathcal{F}$ into the student decoder, and 
obtain output $\hat{\mathcal{B}}$ 
as is in \cref{eq:eq1}. Note that $\mathcal{B}'$ is different from $\hat{\mathcal{B}}$. The former are totally from the teacher and work as a target set, while the latter are predictions which need to be constrained. Moreover, $\mathcal{B}'$ and $\hat{\mathcal{B}}$ are explicitly associated since they all begin 
from the same $\mathcal{Q}'$. Therefore, prediction distillation loss $L_{pd}$ can be directly computed between them.

\setlength{\tabcolsep}{5pt}
\begin{table*}[!h]\footnotesize 
\centering
\begin{tabular}{lcccccccc} 
\toprule
Method  & Backbone & epochs & AP & AP$_{50}$ & AP$_{75}$ & AP$_{S}$ & AP$_{M}$ & AP$_{L}$ \\
\midrule
Def-DETR+ & R50 & 50 & 45.4 & 64.7 & 49.0 & 26.8 & 48.3 & 61.7 \\
Align-DETR & R50 & 50 & 46.0 & 64.9 & 49.5 & 25.2 & 50.5 & 64.7 \\
\rowcolor{gray!20}
OD-DETR (ours) & R50 & 24 & 47.7 (+2.3) & 65.2 & 51.4 & 29.8 & 51.5 & 63.0 \\
\midrule
DAB-Def-DETR* & R50 & 50 & 48.6  & 67.1 & 52.8 & 31.8 & 51.5 & 64.1 \\ 
DN-DAB-Def-DETR & R50 & 50 & 49.4  & 67.5 & 53.8 & 31.2 & 52.5 & 65.1 \\ 
\rowcolor{gray!20}
OD-DAB-DETR (ours) & R50 & 24 & 50.2 (+1.6)  & 67.6 & 54.8 & 33.9 & 54.0 & 65.0 \\ 
\midrule
H-Def-DETR & R50 & 12 & 48.7 & 66.4 & 52.9 & 31.2 & 51.5 & 65.0 \\
DINO-4scale & R50 & 12 & 49.0 & 66.6 & 53.5 & 32.0 & 52.3 & 63.0 \\ 
DINO-4scale & R50 & 24 & 50.4 & 68.3 & 54.8 & 33.3 & 53.7 & 64.8 \\ 
DINO-5scale & R50 & 12 & 49.4 & 66.9 & 53.8 & 32.3 & 52.5 & 63.9 \\ 
Align-DETR & R50 & 12 & 50.2 & 67.8 & 54.4 & 32.9 & 53.3 & 65.0 \\
Align-DETR & R50 & 24 & 51.3 & 68.2 & 56.1 & 35.5 & 55.1 & 65.6 \\
Group-DETR-DINO-4scale & R50 & 36 & 51.3 & - & - & 34.7 & 54.5 & 65.3 \\
Stable-DINO-4scale & R50 & 12 & 50.4 & 67.4 & 55.0 & 32.9 & 54.0 & 65.5 \\ 
Stable-DINO-4scale & R50 & 24 & 51.5 & 68.5 & 56.3 & 35.2 & 54.7 & 66.5 \\ 
Stable-DINO-5scale & R50 & 12 & 50.5 & 66.8 & 55.3 & 32.6 & 54.0 & 65.3 \\ 
\rowcolor{gray!20}
OD-DINO-4scale (ours) & R50 & 12 & 50.4 (+1.4) & 67.4 & 55.2 & 32.9 & 54.1 & 65.6 \\ 
\rowcolor{gray!20}
OD-DINO-4scale (ours) & R50 & 24 & 51.8 (+1.4) & \textbf{69.3} & 56.6 & 34.5 & \textbf{55.1} & \textbf{66.9} \\
\rowcolor{gray!20}
OD-DINO-5scale (ours) & R50 & 12 & \textbf{52.0} (+2.6) & 68.8 & \textbf{56.9} & \textbf{35.4} & 54.8 & 66.8 \\
\bottomrule
\end{tabular}
\caption{
Results for three versions of Our OD-DETR and other detection models using ResNet50 backbone on COCO val2017 dataset. Def-DETR+ indicates Def-DETR with iterative bounding box refinement. DAB-Def-DETR* is the optimized implementation with deformable attention in both encoder and decoder, which is better than the version in 
paper. H-Def-DETR is the model with deformable attention in Hybrid Matching \protect\cite{jia2022detrs}.}
\label{table:all}
\vspace{-0.3cm}
\end{table*}

Here a naive way is to adopt $L_{QFL}$ defined in \cref{eq:eq2} but replace the IoU target $t$ by the class score $c'$ 
from the teacher, which is a strategy for 
distillation under a fixed teacher \cite{Chen2022D3ETRDD}. 
However, the predictions from the EMA teacher 
are not very accurate in our case, especially during the early training stages. 
Applying it in the naive way can introduce many errors from the teacher itself, which 
lead to worse results. We adapt it 
so it becomes suitable for online distillation. First, since $\mathcal{B}'$ have a matching result with GT boxes, an implicit class label index $c_g$ for each box in $\mathcal{B}'$ can be inferred. 
Then we modify the predicted score vector $c'$ at 
the corresponding entry according to \cref{eq:eq6}. 
\begin{equation}\label{eq:eq6}
c'[c_g]\gets(c'[c_g])^\alpha\cdot (IoU')^\beta
\end{equation}
Here $IoU'$ is computed between bounding box in $\mathcal{B}'$ and its matched GT. $\alpha$ and $\beta$ are two hyper-parameters. Following \cite{feng2021tood}, we set $\alpha=0.25$ and $\beta=0.75$. After update on $c'$, we then replace $t$ in \cref{eq:eq2} by it, therefore, giving $L_{distill}^{cls}$ for prediction distillation. For regression task, $c'[c_g]$ is also used as a weight. Using $c'[c_g]$ in $L_{distill}^r$ and $L_{distill}^{cls}$ is named as \textbf{TOOD-weight}, which is first proposed in \cite{feng2021tood} for one-stage detector. Note that $c'[c_g]$ is computed from the teacher, which gives the key difference with \cite{feng2021tood}. Consequently, $L_{distill}^r=c'[c_g](L_1+L_{GIoU})$. $L_{distill}^{cls}$ and $L_{distill}^r$ are combined into $L_{pd}$ in \cref{eq:eq7}.
\begin{equation}\label{eq:eq7}
L_{pd}=\mathds{1}(IoU'>IoU)\cdot L_{distill}^r + L_{distill}^{cls}
\end{equation}
Here $\mathds{1}$ denotes an indicator function, and it only equals to one when the predictions from the teacher have a larger IoU with GT box than the student, otherwise, it returns 0, meaning that $L_{distill}^r$ is not considered in this case. 
This approach is named  as \textbf{Listen2stu}.

\subsection{Auxiliary Group}
To better utilize the EMA teacher and enhance training stability, we choose some updated queries $\tilde{q}$ and corresponding predicted boxes $\tilde{b}$ 
from $(\mathcal{Q}')^t$ and $(\mathcal{B}')^t$ at $t$-th stage of the teacher's decoder. 
Then we use them as independent initial queries and anchors for PE. They are 
fed into the first decoding stage of online model, 
providing more positive examples for learning. To reduce the computational load of the auxiliary group, we select only the top two queries with the lowest matching cost to each GT from each decoder stage. This method ensures that the selected queries include both positive examples and challenging negative ones. Note that queries from the same decoding stage forms an independent group. Predictions from each group is matched with GT set in one-to-one manner. 

Matching distillation is also used in each auxiliary group, just as we do in main group. This method combines the 
original matching from the teacher
with the new ones, which enhances the training stability for auxiliary group. So the loss in auxiliary group can be denoted by $L_{aux}=\tilde{L}_{MQF}+\tilde{L}_{r}$. 
In summary, the total training loss is shown in \cref{eq:eq8}.
\begin{equation}\label{eq:eq8}
L_{total}=L_{MQF}+L_r+L_{pd}+L_{aux}
\end{equation}

\begin{figure}[!h]
    \centering
    \includegraphics[width=7.5cm]{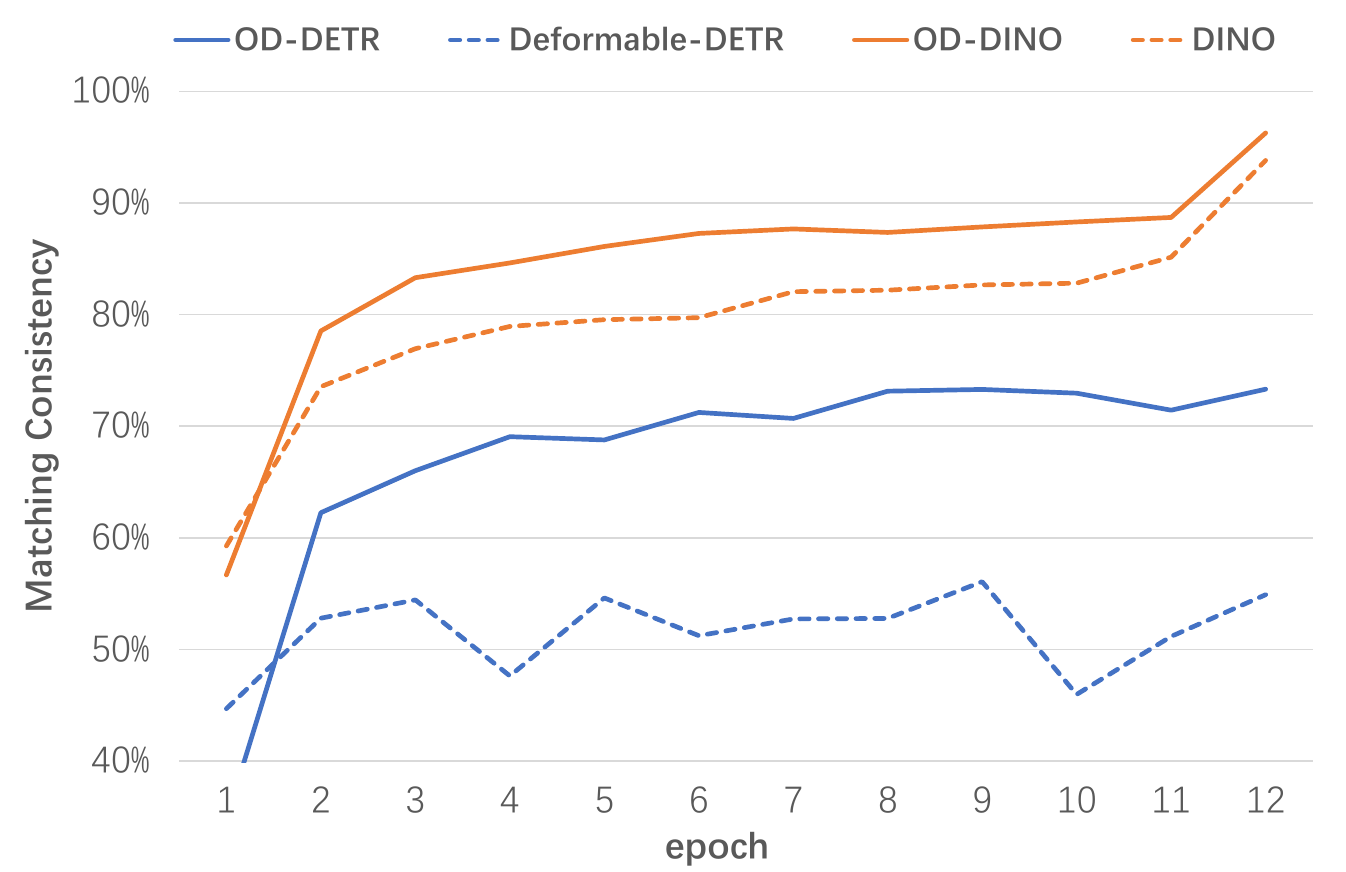}
    \caption{The matching consistency between the EMA and online model. The 
    curve shows the percentage of queries matched with the same GT during training. 
    In both our OD-DETR and OD-DINO models, this ratio is significantly higher compared to Def-DETR and DINO.
    }
    \label{fig:match_consistency}
\end{figure}

\section{Experiments}
\subsection{Settings}

\paragraph{Datasets.} We conduct all our experiments on MS-COCO \cite{lin2014microsoft}
2017 dataset and evaluate the performance of our models on validation dataset by using mean average precision (mAP) metric. The COCO dataset contains 117K training images and 5K validation images.

\paragraph{Implementation details.} We use Def-DETR (with iterative bounding box refinement), DAB-Def-DETR, and DINO as our baseline methods. Our experiments are conducted over 12 (1x) and 24 (2x) epochs on 8 GPUs. Learning rate settings for OD-DETR are identical to those of Def-DETR, with a learning rate of  \( 2 \times 10^{-5} \) for the backbone and \( 2 \times 10^{-4} \) for the Transformer encoder-decoder framework, coupled with a weight decay of \( 2 \times 10^{-5} \). The learning rates and batch sizes for OD-DAB-DETR and OD-DINO follow their respective baselines. We set the EMA decay value at 0.9996.

\subsection{Main Results}
As shown in Table \ref{table:all}, we firstly use Def-DETR \cite{zhu2020deformable} as baseline, and compare our OD-DETR on COCO val2017 dataset with other competitive DETR variants. 
Our OD-DETR achieved a significant improvement, reaching 47.7 AP. It notably surpassed the baseline (Def-DETR iterative bounding box refinement) trained for 50 epochs, with a substantial increase of 2.3 in AP after just 2x training.

The OD-DAB-DETR model, improving upon the DAB-Def-DETR \cite{liu2022dab}, achieved 50.2 AP. It exceeds the baseline by 1.6 AP after just 2x training and outperforms DN-Def-DETR \cite{li2022dn} with 50 epochs of training.

Under both 1x and 2x schedulers, our OD-DINO consistently outperforms the baseline DINO \cite{zhang2022dino}, achieving 50.4 AP with 1x and 51.8 AP with 2x scheduling, marking a 1.4 increase in AP.
Our OD-DINO series outperform other DINO-based approaches, including Stable-DINO \cite{liu2023detection}, Align-DETR \cite{cai2023aligndetr} and so on.

\setlength{\tabcolsep}{4pt}
\begin{table}[ht]\footnotesize 
\centering
\begin{tabular}{l|ccc|ccc} 
\toprule
Method  & MD & PD & AG & AP & AP$_{50}$ & AP$_{75}$ \\
\midrule
0 (baseline) &  &  &  & 45.4 & 64.7 & 49.0 \\
\midrule
1 & \checkmark &  &  & 46.8 & 64.6 & 50.7 \\
2 & \checkmark & \checkmark &  & 47.3 & 65.0 & 51.4 \\
3 & \checkmark &  & \checkmark & 47.2 & 65.0 & 51.2 \\
4 & \checkmark & \checkmark & \checkmark & \textbf{47.7} & \textbf{65.2} & \textbf{51.4} \\
\bottomrule
\end{tabular}
\caption{
Ablations on all designed components. MD stands for matching distillation. PD refers to prediction distillation. AG means auxiliary group 
built by updated queries and boxes. 
}
\label{table:ablation}
\end{table}

\subsection{Ablation Study}
We conduct a series of ablation studies to assess the effectiveness of the components. All experiments in this section are based on the Def-DETR model with an R50 backbone and follow a standard 2x training schedule.

\subsubsection{Ablation Study on Components}
In 
this section, we perform a set of ablation studies on key components: matching distillation (MD), prediction distillation (PD), and auxiliary group (AG), to assess their individual effectiveness. The findings are detailed in Table \ref{table:ablation}. These studies reveal that each component plays a significant role in enhancing performance. Specifically, MD contributes an increase of 1.4 AP, PD adds 0.5 AP, and AG contributes 0.4 AP. Cumulatively, these components lead to an overall improvement of 2.3 AP over the baseline model.

Fig.\ref{fig:match_consistency} shows that our methods improves the consistency of matching GTs between the online model's queries and the EMA model's queries. As the online model's matching improves with training, it also boosts the EMA model's stability. This two-way improvement creates a strong combined effect.

\setlength{\tabcolsep}{4pt}
\begin{table}[!ht]\footnotesize 
\centering
\begin{tabular}{l|ccc}
\toprule
Method & AP & AP$_{50}$ & AP$_{75}$ \\
\midrule
QFL w/o MD  & 45.8 & 63.6 & 49.8 \\
\midrule
Conditional MD & 45.9 & 64.1 & 49.7 \\
QFL MD & \textbf{46.4} & \textbf{64.1} & \textbf{50.5} \\
QFL + Conditional MD & 46.3 & 64.1 & 49.9 \\
\bottomrule
\end{tabular}
\caption{Comparison on 
various matching distillation methods 
for classification. QFL w/o MD means only applying QFL without MD. QFL MD refers to our MD combined with $L_{MQF}$. Condition MD is introduced in \protect\cite{Teng2023StageInteractorQO}.}
\label{table:MDC_type}
\end{table}

\subsubsection{Comparisons of Different Matching Distillation Methods}
Experiments in Table \ref{table:MDC_type} focuses on different matching distillation methods for only classification. The first row shows that employing only QFL \cite{li2020generalized} without MD results in a significantly lower performance, achieving 45.8 AP, compared to our MD method combined with $L_{MQF}$ (QFL MD), which reaches 46.4 AP. This difference highlights the significant contribution of MD beyond the impact of QFL alone.
Conditional MD, introduced in the StageInteractor \cite{Teng2023StageInteractorQO}, is a technique that allows a query to match with multiple ground truths. This method involves verifying if the additional query-GT matches have an IoU above 0.5, discarding those that don't meet this criterion. The performance of it is 45.9 AP.
Integrating $L_{MQF}$ with Conditional MD under the label QFL + Conditional MD doesn't enhance performance. The performance is 46.3 AP. This observation implies that $L_{MQF}$ and MD are inherently synergistic. $L_{MQF}$ eliminates the necessity of manually setting thresholds like in Conditional MD for Hungarian matching.

\setlength{\tabcolsep}{4pt}
\begin{table}[!htb]\footnotesize 
\centering
\begin{tabular}{l|c|c|ccc}
\toprule
Method & Cls & Reg & AP & AP$_{50}$ & AP$_{75}$ \\
\midrule
1 & w/o $w_d$ & N/A & 46.4 & 64.1 & 50.5 \\
\midrule
2 & w/o $w_d$ & w/o $w_d$ & 46.6 & 64.5 & 50.6 \\
3 & w/o $w_d$ & w/ $w_d$ & \textbf{46.8} & \textbf{64.6} & \textbf{50.7} \\
4 & w/ $w_d$ & w/ $w_d$ & 46.7 & 64.3 & 50.5 \\
\bottomrule
\end{tabular}
\caption{Analyzing the impact of using down-weight ($w_d$) in MD. N/A in the first line indicates that MD was not applied for regression.}
\label{table:MDR_type}
\end{table}

\begin{figure}[htpb]
    \centering
    \includegraphics[width=7.5cm]{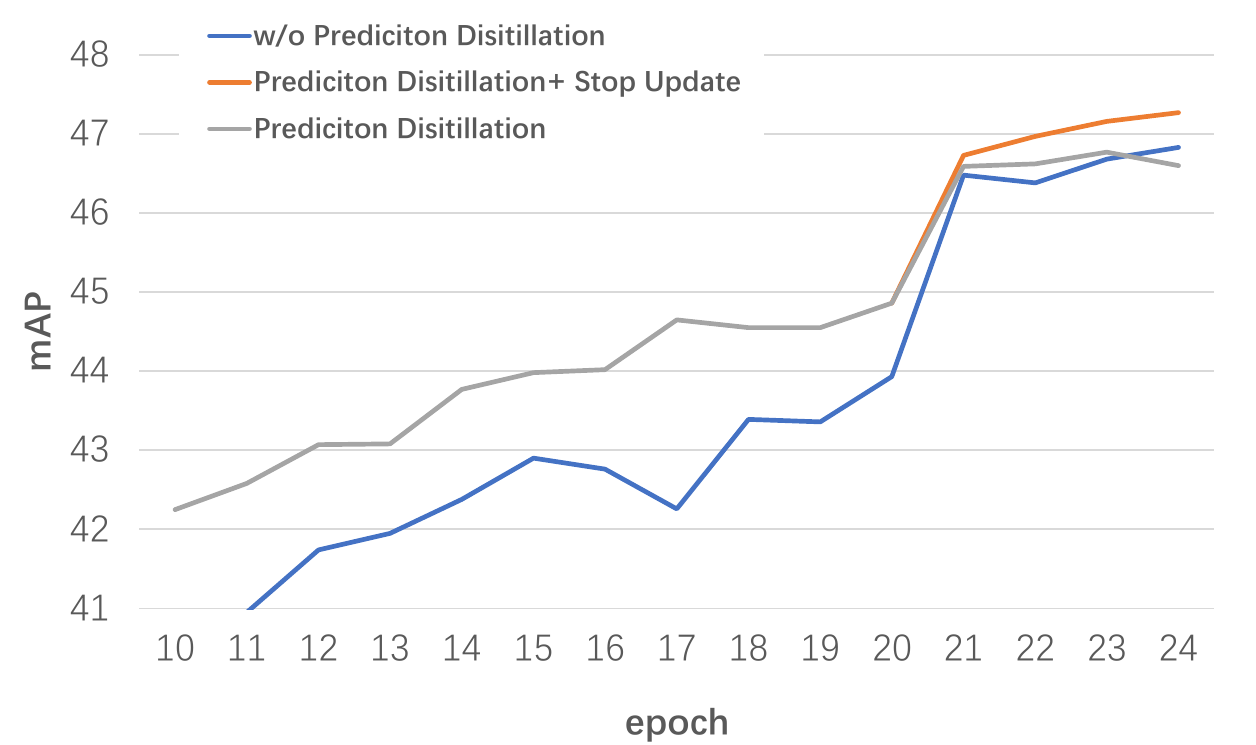}
    \caption{The Stop Update strategy extends the advantage of PD from before the learning rate decay to after it. This ensures that PD ultimately achieves better performance than methods without PD.}
    \label{fig:stop_update}
\end{figure}

\subsubsection{Analyzing Down-Weight Impact in Matching Distillation}
In Method 1 from Table \ref{table:MDR_type}, MD is only used for classification, keeping a one-to-one match for regression, resulting in a lower AP of 46.4 (QFL MD in Table \ref{table:MDC_type}). This implies that MD is beneficial for regression. 
Method 3, which applies down-weight ($w_d$) only for regression, achieves the highest AP of 46.8. This confirms the effectiveness of down-weighting regression loss. It also demonstrates QFL MD's effectiveness in handling many-to-many matching in classification. It eliminates the need to adjust loss for GTs that match multiple queries.

\subsubsection{Comparisons of Various Prediction Distillation Methods}
In Table \ref{table:distill_type}, w/o PD means using MD alone without PD, and it's also referred to Method 1 in Table \ref{table:ablation}. Naive distillation applies distillation evenly across all queries, but this method only achieves 44.2 AP, lower than w/o PD (46.8 AP). This suggests that simply adding online distillation doesn't work.
Using TOOD-weight and Listen2stu methods led to a notable increase of 2.3 AP, reaching 46.6 AP. As Fig.\ref{fig:stop_update} shows, PD  notably enhances performance before learning rate decay, but tends to have a negative impact after the learning rate decay. Thus, we adopt a \textbf{Stop Update} strategy, halting updates to the EMA teacher's weights after learning rate decay. This method eventually achieves 47.3 AP.
We also test Query-prior Assignment Distillation from DETRDistill \cite{Chang2022DETRDistillAU}. It feeds the teacher's query into the student's decoder as an extra group to learn GT. The outcome, at 46.9 AP, is less effective than our PD approach.

\setlength{\tabcolsep}{4pt}
\begin{table}[!htb]\footnotesize 
\centering
\begin{tabular}{l|ccc}
\toprule
Method & AP & AP$_{50}$ & AP$_{75}$ \\
\midrule
w/o PD & 46.8 & 64.6 & 50.7 \\
\midrule
Naive Distillation & 44.2 & 62.6 & 47.7 \\
TOOD-weight & 46.5 & 64.4 & 50.4 \\
TOOD-weight + Listen2stu & 46.6 & 64.4 & 50.6 \\
TOOD-weight + Listen2stu + Stop Update  & \textbf{47.3} & \textbf{65.0} & \textbf{51.4} \\
\midrule
 Query-prior Assignment Distillation & 46.9 & 64.5 & 50.9 \\
\bottomrule
\end{tabular}
\caption{Comparisons of various methods of PD. 
w/o PD means that only MD is used, and PD is not utilized. 
Stop Update refers to ceasing the update of the teacher model after learning rate decay.}
\label{table:distill_type}
\end{table}

\setlength{\tabcolsep}{4pt}
\begin{table}[!h]\footnotesize 
\centering
\begin{tabular}{l|ccc}
\toprule
Method & AP & AP$_{50}$ & AP$_{75}$ \\
\midrule
Original Matching & 46.7 & 64.1 & 50.8 \\
Re-Matching & 47.0 & 64.5 & 50.9 \\
MD & \textbf{47.2} & \textbf{65.0} & \textbf{51.2} \\
\bottomrule
\end{tabular}
\caption{Experiments on applying MD to the auxiliary group.  Original Matching uses the original matching results from the EMA model. Re-Matching involves utilizing a new Hungarian matching results during the training of the auxiliary group. MD combines the two match results.}
\label{table:aux_grp_method}
\end{table}

\subsubsection{Experiments on Applying MD to the Auxiliary Group}
Table \ref{table:aux_grp_method} shows results for different matching methods in auxiliary group. Using either the EMA model's original matching or re-matching gives 47.0 AP and 46.7 AP, respectively. The highest AP, 47.2, comes from using MD for auxiliary group (Method 3 in Table \ref{table:ablation}), just like in the main group. This proves MD works well in both main and auxiliary group.

\subsubsection{Comparative Results of EMA Models}
Table \ref{table:ema_rt} compares the EMA model performances of DINO-4scale and our OD-DINO-4scale, both trained using a 2x setting. Our OD-DETR-EMA outperforms DINO-EMA by 1.3 AP. DINO's EMA version achieves 50.7 AP, which is lower than the the online version of OD-DINO (51.8 AP). This indicates that our methods not only improves the performance of the online student but also enhances the EMA teacher itself. A better teacher then further improves the student, creating a beneficial cycle.

\setlength{\tabcolsep}{5pt}
\begin{table}[htpb]\footnotesize 
\centering
\begin{tabular}{l|ccc}
\toprule
Method & AP & AP$_{50}$ & AP$_{75}$ \\
\midrule
DINO-EMA & 50.7 & 68.4 & 55.2 \\
OD-DINO-EMA & \textbf{52.0} (+1.3) & \textbf{69.4} & \textbf{56.8} \\
\bottomrule
\end{tabular}
\caption{Performance comparison of EMA models. }
\label{table:ema_rt}
\end{table}

\section{Conclusion}
This paper proposes an OD-DETR, which is an online distillation method to stabilize the training of DETR. We find that the EMA model accumulated during training provides not only high-quality predicted boxes, but also the query-GT matching result and extra query groups. Under the help of EMA model, we improve the online training through prediction distillation, matching distillation and auxiliary query group. We show that the proposed OD-DETR can steadily increase the performance across different DETR variations. 

\section*{Acknowledgments}
This work is supported by the Science and Technology Commission of Shanghai Municipality under Grant No. 22511105800, 19511120800 and
22DZ2229004.

\bibliographystyle{named}
\bibliography{ijcai24}

\clearpage
\appendix

\section{Visualizations of Decoder Sampling Points}
\begin{figure*}[h]
    \vspace{-0.2cm}
    \centering
    \includegraphics[width=16.5cm]{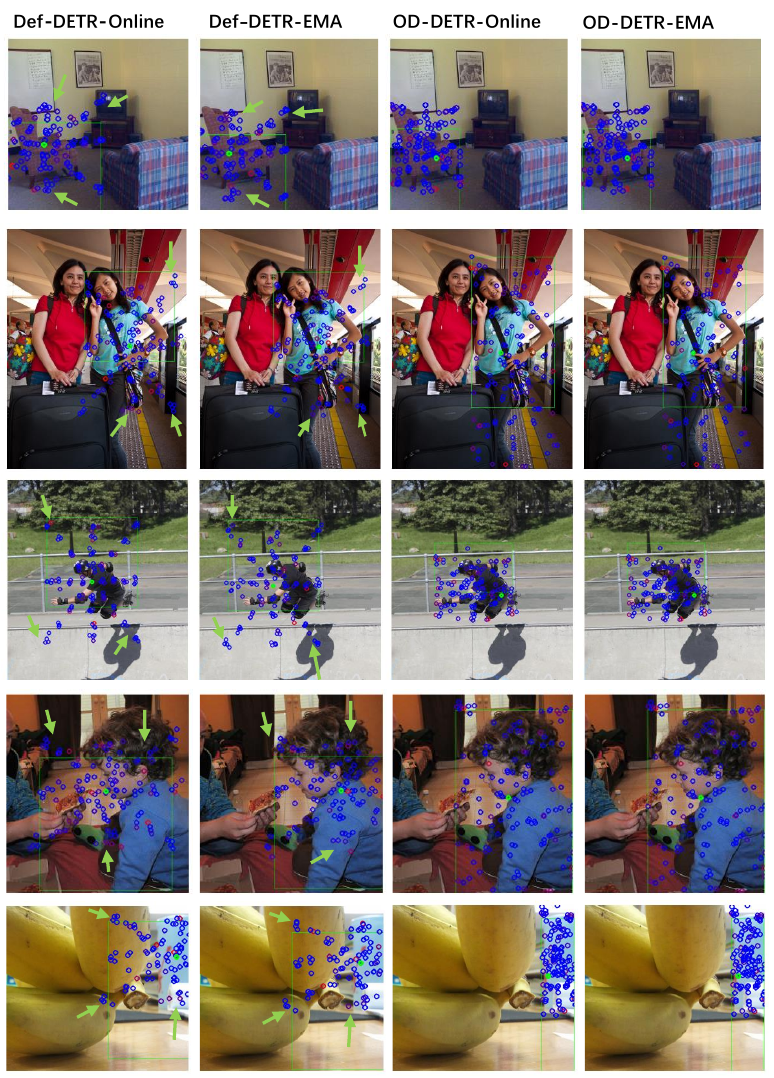}
    \caption{Visualization of multi-scale deformable cross attention: Each sampling point is represented as a circle, with the color indicating the corresponding attention weight. A redder circle denotes a higher weight, while a bluer circle signifies a lower weight. The reference point is marked with a green circle, and the predicted bounding box is outlined as a green rectangle. The green arrows in the figure point to areas where there are significant differences in the sampling points between the online Def-DETR and its EMA model.}
    \label{fig:sampling}
\end{figure*}

To assess the effectiveness of online distillation, we compare the sampling points of Def-DETR \cite{zhu2020deformable} and OD-DETR with their respective EMA models in Fig. \ref{fig:sampling}. All models are conducted over 12 (1x) training. The first and second columns show the results of Def-DETR and its EMA version, while the third and fourth columns present the results of OD-DETR and its EMA version, respectively.

First, it's clear that the sampling points of OD-DETR better cover the foreground object areas, and the resulting prediction boxes are more accurate than those of Def-DETR. This reaffirms the effectiveness of our method. Second, the prediction boxes, reference points, and sampling points produced by OD-DETR and its EMA version are nearly identical. In contrast, there are significant differences in these elements between Def-DETR and its EMA version. This also confirms that our online distillation method effectively enables the online student to learn from the EMA teacher.

\section{Sharing Decoder Parameters Among Stages}
Inspired by \cite{Zhao2023RecursiveDetER}, we 
share decoder 
parameters among different stages in Def-DETR \cite{zhu2020deformable}. This approach is more compatible with our online distillation method. As shown in Table \ref{table:share_dec}, when Def-DETR shares its decoder, its AP decreases from the original 45.9 to 45.8. Conversely, our OD-DETR increases its AP from 47.2 to 47.7. Consequently, we choose to share decoder parameters in OD-DETR. This demonstrates that our method not only increases the stability of GT-query matching during training, but also enhances performance while reducing the number of model parameters.

\setlength{\tabcolsep}{4pt}
\begin{table}[!h]\footnotesize 
\centering
\vspace{-0.1cm}
\begin{tabular}{c|c|c|ccc} 
\toprule
Method & Params & Share-decoder & AP & AP50 & AP75 \\
\midrule
Def-DETR & 406M &  & 45.9 & 64.9 & 49.3 \\
Def-DETR & 347M & \checkmark & 45.8 & 64.1 & 49.3 \\
\midrule
OD-DETR & 406M &  & 47.2 & 65.5 & 51.0 \\
OD-DETR & 347M & \checkmark & 47.7 & 65.2 & 51.4 \\
\bottomrule
\end{tabular}
\caption{Comparison of Applying Shared Decoder to Def-DETR and OD-DETR}
\label{table:share_dec}
\vspace{-0.2cm}
\end{table}

For DINO \cite{zhang2022dino}, parameters of 
classification and regression heads at each stage are already shared. As shown in Table \ref{table:share_dec_dino}, we also conduct experiment 
by sharing other decoder parameters. 
However, unlike in Def-DETR, 
this efficient scheme 
leads to a decrease in AP from 50.4 to 49.9. We speculate 
the degradation could be due to the Look Forward Twice method 
in DINO, which allows the gradient flow of regression loss 
from the current stage 
the previous one, 
and becomes redundant and unstable in the shared decoder. 
Therefore, we retain the setting in original DINO without 
sharing decoder parameters. 

\setlength{\tabcolsep}{4pt}
\begin{table}[!h]\footnotesize 
\centering
\vspace{-0.1cm}
\begin{tabular}{c|c|ccc} 
\toprule
Method & Share-decoder & AP & AP50 & AP75 \\
\midrule
OD-DINO-4scale &  & 50.4 & 67.4 & 55.2 \\
OD-DINO-4scale & \checkmark & 49.9 & 67.1 & 54.7 \\
\bottomrule
\end{tabular}
\caption{Comparison of Applying Shared Decoder to OD-DINO}
\label{table:share_dec_dino}
\vspace{-0.2cm}
\end{table}

\section{Performances of COCO val2017 with Swin-L backbone}
We also conduct experiments using Swin-L as the backbone. The results shown in Table \ref{table:swin} further confirm OD-DETR's effectiveness: it enhances the DINO baseline by 0.9 AP in just 12 epochs. Ultimately, the final model achieves a performance of 57.7 AP.

\section{Exeriments on COCO test-dev}
We validate our OD-DETR series on COCO test-dev in Table \ref{table:test_dev}. The performance of all our models on both test-dev and val2017 is quite similar, and each shows a significant improvement over the baseline, indicating the robust performance of our models.

\setlength{\tabcolsep}{4pt}
\begin{table}[!h]\footnotesize 
\centering
\vspace{-0.1cm}
\begin{tabular}{c|c|c|ccc} 
\toprule
Method & Epochs & Backbone & AP & AP50 & AP75 \\
\midrule
OD-DETR & 24 & R50 & 47.8 & 65.5 & 52.0 \\
OD-DAB-DETR & 24 & R50 & 50.2 & 67.8 & 54.7 \\
\midrule
OD-DINO-4scale & 12 & R50 & 49.9 & 66.9 & 54.7 \\
OD-DINO-4scale & 24 & R50 & 51.6 & 68.9 & 56.3 \\
OD-DINO-5scale & 12 & R50 & 51.9 & 68.7 & 56.9 \\
\midrule
OD-DINO-4scale & 12 & Swin-L & 57.8 & 76.0 & 63.2 \\
\bottomrule
\end{tabular}
\caption{Performance of OD-DETR Series on COCO test-dev}
\label{table:test_dev}
\vspace{-0.2cm}
\end{table}

\setlength{\tabcolsep}{2pt}
\begin{table}[h]\scriptsize 
\centering
\vspace{0 cm}
\begin{tabular}{ccccccccc} 
\toprule
Method  & Backbone & Epochs & AP & AP$_{50}$ & AP$_{75}$ & AP$_{S}$ & AP$_{M}$ & AP$_{L}$ \\
\midrule
H-DETR & Swin-L & 12 & 56.1 & 75.2 & 61.3 &  39.3 & 60.4 & 72.4 \\
H-DETR & Swin-L & 36 & 57.6 & 76.5 & 63.2 & 41.4 & 61.7 &  73.9 \\
Co-DETR & Swin-L & 12 & 49.0 & 66.6 & 53.5 & 32.0 & 52.3 & 63.0 \\
DINO-4scale & Swin-L & 12 & 56.8 & 75.6 & 62.0 & 40.0 & 60.5 & 73.2 \\ 
DINO-4scale & Swin-L & 36 & 58.0 & 77.1 & 66.3 & 41.3 & 62.1 & 73.6 \\ 
Stable-DINO-4scale & Swin-L & 12 & 57.7 & 75.7 & 63.4 & 39.8 & 62.0 & 74.7 \\
Stable-DINO-4scale & Swin-L & 24 & 58.6 & 76.7 & 64.1 & 41.8 & 63.0 & 74.7 \\
\midrule
\rowcolor{gray!20}
OD-DINO-4scale & Swin-L & 12 & 57.7 (+0.9) & 75.8 & 63.2 & 41.1 & 61.8 & 74.8 \\ 
\rowcolor{gray!20}
OD-DINO-4scale & Swin-L & 24 & \textbf{58.8} (+0.8) & \textbf{77.0} & \textbf{64.4} & \textbf{42.4} & \textbf{63.1} & \textbf{75.1} \\ 
\bottomrule
\end{tabular}
\caption{Comparison of previous DETR variants on COCO val2017 using Swin-L backbones}
\label{table:swin}
\end{table}

\end{document}